\def\year{2022}\relax
\documentclass[letterpaper]{article} % DO NOT CHANGE THIS
\usepackage{aaai22}  % DO NOT CHANGE THIS
\usepackage{times}  % DO NOT CHANGE THIS
\usepackage{helvet}  % DO NOT CHANGE THIS
\usepackage{courier}  % DO NOT CHANGE THIS
\usepackage[hyphens]{url}  % DO NOT CHANGE THIS
\usepackage{graphicx} % DO NOT CHANGE THIS
\usepackage{natbib}  % DO NOT CHANGE THIS AND DO NOT ADD ANY OPTIONS TO IT
\usepackage{caption} % DO NOT CHANGE THIS AND DO NOT ADD ANY OPTIONS TO IT
\DeclareCaptionStyle{ruled}{labelfont=normalfont,labelsep=colon,strut=off} % DO NOT CHANGE THIS
\usepackage{algorithm}
\usepackage{algorithmic}

\usepackage{newfloat}
\usepackage{listings}
\floatstyle{ruled}
\newfloat{listing}{tb}{lst}{}
\floatname{listing}{Listing}

\usepackage{latexsym}
\usepackage{subfigure}
\usepackage{makecell}
\usepackage{bm}
\usepackage{amssymb}
\usepackage{amsmath}
\usepackage{dblfloatfix}
\usepackage{enumitem}
\setlist{leftmargin=*}%, itemsep=0pt}
\usepackage{comment}
\usepackage{multicol}
\usepackage{multirow}

\title{SAS: Self-Augmentation Strategy for Language Model Pre-training}
\author {
    Yifei Xu\textsuperscript{\rm 1}\equalcontrib, Jingqiao Zhang\textsuperscript{\rm 2}\equalcontrib, Ru He\textsuperscript{\rm 2}\equalcontrib, Liangzhu Ge\textsuperscript{\rm 2}\equalcontrib, 
  \textbf{Chao Yang\textsuperscript{\rm 2}, Cheng Yang\textsuperscript{\rm 2}\thanks{Corresponding author: Cheng Yang.}, Ying Nian Wu\textsuperscript{\rm 1}}
}
\affiliations {
    \textsuperscript{\rm 1} University of California, Los Angeles \\
    \textsuperscript{\rm 2} Alibaba Group\\
  fei960922@ucla.edu, \{jingqiao.zhang, ru.he, liangzhu.glz, xiuxin.yc, charis.yangc\}@alibaba-inc.com, ywu@stat.ucla.edu \\  
}

\begin{document}

\maketitle

\begin{abstract}
The core of self-supervised learning for pre-training language models includes pre-training task design as well as appropriate data augmentation. 
Most data augmentations in language model pre-training are context-independent. A seminal contextualized augmentation was recently proposed in ELECTRA and achieved state-of-the-art performance by introducing an auxiliary generation network (generator) to produce contextualized data augmentation for the training of a main discrimination network (discriminator). This design, however, introduces extra computation cost of the generator and a need to adjust the relative capability between the generator and the discriminator. In this paper, we propose a self-augmentation strategy (SAS) where a single network is utilized for both regular pre-training and contextualized data augmentation for the training in later epochs. Essentially, this strategy eliminates a separate generator and uses the single network to jointly conduct two pre-training tasks with MLM (Masked Language Modeling) and RTD (Replaced Token Detection) heads. 
It avoids the challenge to search for an appropriate size of the generator, 
which is critical to the performance as evidenced in ELECTRA and its subsequent variant models. 
In addition, SAS is a general strategy that can be seamlessly combined with many new techniques emerging recently or in the future, such as the disentangled attention mechanism from DeBERTa. Our experiments show that SAS outperforms ELECTRA and other state-of-the-art models in the GLUE tasks with similar or less computation cost.
\end{abstract}

\section{Introduction} \label{sec_intro}
In recent years, language model pre-training has achieved enormous success in various natural language processing (NLP) downstream tasks,
largely due to innovations in pre-training task design as well as associated data augmentation. 
Especially, the BERT model \citep{bert_devlin2018} is a milestone, where each input token sequence is augmented by randomly masking a proportion (typically 15\%) of tokens
and then a masked language model (MLM) task is undertook by recovering each masked token out of the vocabulary from the augmented input.
Since its success, BERT has been refined or extended in the literature by adding or varying pre-training tasks and/or data augmentation.
For example, 
the RoBERTa \citep{RoBERTa_Liu_2019} empirically shows that dynamically masking tokens boosts BERT's performance,
while pairing  %irrelevant 
sentences and undertaking a corresponding next sentence prediction (NSP) task does not;
the SpanBERT \citep{SpanBERT_Joshi_2019} masks spans of tokens in each input sequence and then uses two boundary tokens of each masked span to predict each token in the span;
the StructBERT \citep{StructBERT_Wang_2019} shuffles a proportion of unmasked tri-grams in each input sequence and then re-constructs the right order of these tokens;
the XLNet \citep{XLNet_Yang_2019} permutes 
the token index order in each input sequence and then autoregressively predicts the ending tokens in each given order;
the UNILM \citep{UNILM_Dong_NIPS2019} employs different self-attention masks to augment input data in three ways so that unidirectional, bidirectional and sequence-to-sequence LM tasks can be performed together;
the ERNIE \citep{ERNIE_Sun_2019} uses prior knowledge to enhance its data augmentation by phrase-level masking and entity-level masking;
the CLEAR \citep{clear_wu2020}, while keeping the MLM task, generates a pair of augmented sentences from each input sentence using four augmentation methods 
(words deletion, spans deletion, synonym substitution and reordering) 
and then performs a sentence-level contrastive learning task to minimize the distance between augmented sentences coming from the same original sentence.
In all these models, data augmentations are manually designed and context-independent without taking sequence context into account.

Recently, the ELECTRA \citep{electra_clark2020} model is proposed to construct contextualized data augmentation. 
It trains an auxiliary generation network to provide plausible tokens to replace a proportion (typically 15\%) of the original tokens according to the input context, and then utilize a main discrimination network to classify whether each token in the augmented input is a replaced token or not. The latter task was named as replaced token detection (RTD). 
After pre-training, the generator is discarded and the discriminator is further fine-tuned for downstream NLP tasks.
ELECTRA has shown impressive advantages over MLM-based models in various downstream tasks under similar computation cost,
especially when a model size is small.
Since the ELECTRA model needs to train two separate neural networks (i.e., generator and discriminator) side by side, 
however,
it is challenging to maintain an appropriate level of generator's ability relative to discriminator's during the training.
When the generator becomes 
stronger than what the discriminator is capable of dealing with,
it will prevent effective learning of the discriminator. In addition, training two separate networks naturally leads to extra computation cost.

To address the aforementioned disadvantages, we propose a self-augmentation strategy (SAS)
that uses a single neural network to provide contextualized data augmentation for its own use in subsequent epochs during the training.\footnote{
In this paper, we focus on the field of language modeling and note that SAS is a general strategy that can also be applied to other domains such as computer vision and pattern recognition.} The SAS network includes two light-weight heads (MLM head and RTD head) on the top of one common transformer encoder \citep{Transformer_NIPS2017}.
This network architecture allows automatic adjustment of the relative difficulty of the generated tokens from the MLM head, while jointly undertaking both MLM and RTD pre-training task.
As a result, our SAS can achieve higher downstream GLUE scores   \citep{GLUE_Wang_2018} 
than ELECTRA using the same or less computation cost.
The advantages of SAS can be summarized as follows:

\begin{itemize}
\item Compared to ELECTRA and its subsequent variants that utilize an auxiliary network to generate contextualized data augmentation, 
SAS introduces a single network to jointly generate contextualized data augmentation and perform pre-training tasks. On one hand, it avoids the challenge to balance the capability of the generator and that of other model components. 
On the other hand, it has both MLM and RTD heads on the top of the same encoder, 
so that the pre-trained SAS model have a foundation for both generation and discrimination capabilities 
which is beneficial to the fine-tuning of different downstream tasks. 
As a comparison, in the case of ELECTRA, the generation network is discarded and only the discrimination network is used for downstream applications.

\item SAS eliminates the computation cost related to the generation network and therefore further improves computation efficiency. 
While ELECTRA has already shown its computation efficiency (especially for small models), 
SAS provides an extra economic pre-training solution for many scenarios where compute resource is limited .

\item 
SAS is a simple and general strategy so that it is able to easily incorporate many other techniques emerging recently or in the future.
For example, 
when SAS incorporates the disentangled attention mechanism proposed in DeBERTa \cite{deberta_he2020},
it can significantly outperform
the previous state-of-the-art models in the GLUE tasks using the same or less computation cost.

\end{itemize}

\section{Related Work} \label{sec_related_work}

\textbf{Pre-trained Language Models:}
As introduced in 
the previous section, 
many pre-trained language models have emerged with new performance records in many downstream NLP tasks in the last several years. 
Among them, ELECTRA \citep{electra_clark2020} distinguishes itself from previous models and inspires our SAS model,
since it introduces contextualized data augmentation with an auxiliary generation network and then train a main discrimination network to perform the RTD task.

The Electric model is later proposed by \citet{Electric_Clark_EMNLP2020} as an energy-based model to perform the cloze task \citep{Cloze_Taylor_JQ1953} using noise-contrastive estimation \citep{NCE_Gutmann_JMLR2013}. 
It is particularly effective at producing likelihood scores for text but slightly under-performs ELECTRA on the GLUE tasks.
This model also needs a separate network
to provide noise distribution in order to generate contextualized data augmentation. The MC-BERT model is proposed by \citet{MC-BERT_Xu_2020} to replace the RTD binary classification task in ELECTRA
with a multi-class classification task so that richer semantic information can be captured.
While MC-BERT performs on par with ELECTRA,
it also needs a separate network called meta controller to generate a multi-class candidate set for each token as contextualized data augmentation. 

COCO-LM \citep{coco_meng2021} has been recently proposed to improve ELECTRA by 
using two new pre-training tasks called Corrective Language Modelling (CLM) and Sequence Contrastive Learning tasks.
In particular, the CLM refines the All-Token MLM task studied by \citet{electra_clark2020} 
and jointly learns the RTD and MLM tasks by a combination of two loss objectives at all output token positions. This is similar to SAS; however,  a key caveat is that the MLM head, which is computationally much heavier than the RTD head due to its softmax calculation over the vocabulary, is only conducted at the augmented positions in SAS.
COCO-LM also relies on a separate generation network to generate the augmented sequence for its two tasks 
which leads to extra computation.
In contrast, the proposed SAS eliminates the requirement of a separate generator and achieves higher computation efficiency.

Another recent work is the DeBERTa model \citep{deberta_he2020}, 
which computes the attention weights among tokens using disentangled matrices on two separate vectors (content vector and relative position vector) of each token.
Because SAS is a general strategy, it can seamlessly incorporate such a disentangled attention mechanism to further improve performance, as will be shown in 
Section of Experiments.

\textbf{Teacher-student Mechanism:} 
SAS can be alternatively viewed from the perspective of a teacher-student mechanism, 
where the knowledge of a teacher model is utilized to facilitate the learning process of a student model.
In essence, in the SAS method, we treat the model trained in the current epoch as a student and that in the previous epoch as a (weak) teacher, since the latter generates contextualized data augmentation to help the former to learn.

The teacher-student mechanism has been traditionally used in knowledge distillation for the purpose of label augmentation \citep{KD_Hinton_NIPS_2015},
where soft labels generated from a teacher is combined with hard (one-hot) labels observed from the original data 
to serve as the ultimate label to supervise the learning of a student.
The size of a student could be either smaller than or the same as that of a teacher.  
\citet{DistilBERT_Sanh_2019} shows that using the original BERT as a teacher
their DistilBERT (student) model is able to remain 97\% of the teacher's performance in the downstream GLUE tasks, 
while the size of the student is only 60\% of that of the teacher.
When a student has the same size as its teacher during the knowledge distillation, 
the advantages of student over teacher have been empirically shown in recent research.
For example,
\citet{Self-KD_Kim_2020} show the benefits of their Teacher-free Knowledge Distillation method by using a pre-trained first-generation model as a teacher,
while \citet{Revisiting_KD_Yuan_CVPR_2020} show the gains of their Self-knowledge Distillation method by using a (student) model itself in the previous epoch during the training as a teacher.

In recent years, the teacher-student mechanism has also been used to obtain state-of-the-art results in semi-supervised settings. 
\citet{Temporal_Ensembling_Laine_2016} propose Temporal Ensembling method to use 
an EMA (exponential moving average) of the output predictions of the model in the previous epochs as a teacher prediction
to constitute the consistency loss component for each data instance, 
in addition to the cross-entropy loss component for each labeled data instance.
\citet{Mean_Teachers_Tarvainen_NIPS_2017} propose to use an EMA of parameter values (i.e., weights) of (student) model
to construct a mean teacher for the consistency loss component for each data instance.

\textbf{Curriculum Learning and Self-paced Learning}
Curriculum learning and self-paced learning have attracted increasing attention in machine learning. 
Both of them generally claim that it helps the learning process to start learning with easier data and then gradually take more complex or difficult data into consideration.
There are some notable works that intend to use them in various NLP tasks ~\citep{self_wan_2020,curriculum_xu_2020, nijkamp2021script}. 
In SAS, we naturally provides a curriculum for the model itself by incrementally augmenting the data as MLM improves generation capability over the training process. 
In essence, our model can be regarded as a combination of the two learning strategies which adaptively generate augmented data for better learning. 
This kind of idea dates back to the work of \citet{self_jiang_aaai2015}, 
which is a pioneer of combining these two strategies.
\section{Method} \label{sec_method}
In this section, we formulate the self-augmentation strategy as a general framework that can be applied to various domains, and then detail its usage in language model pre-training.

\subsection{General SAS Framework}

\begin{figure}
	\centering 
	\includegraphics[width=\linewidth]{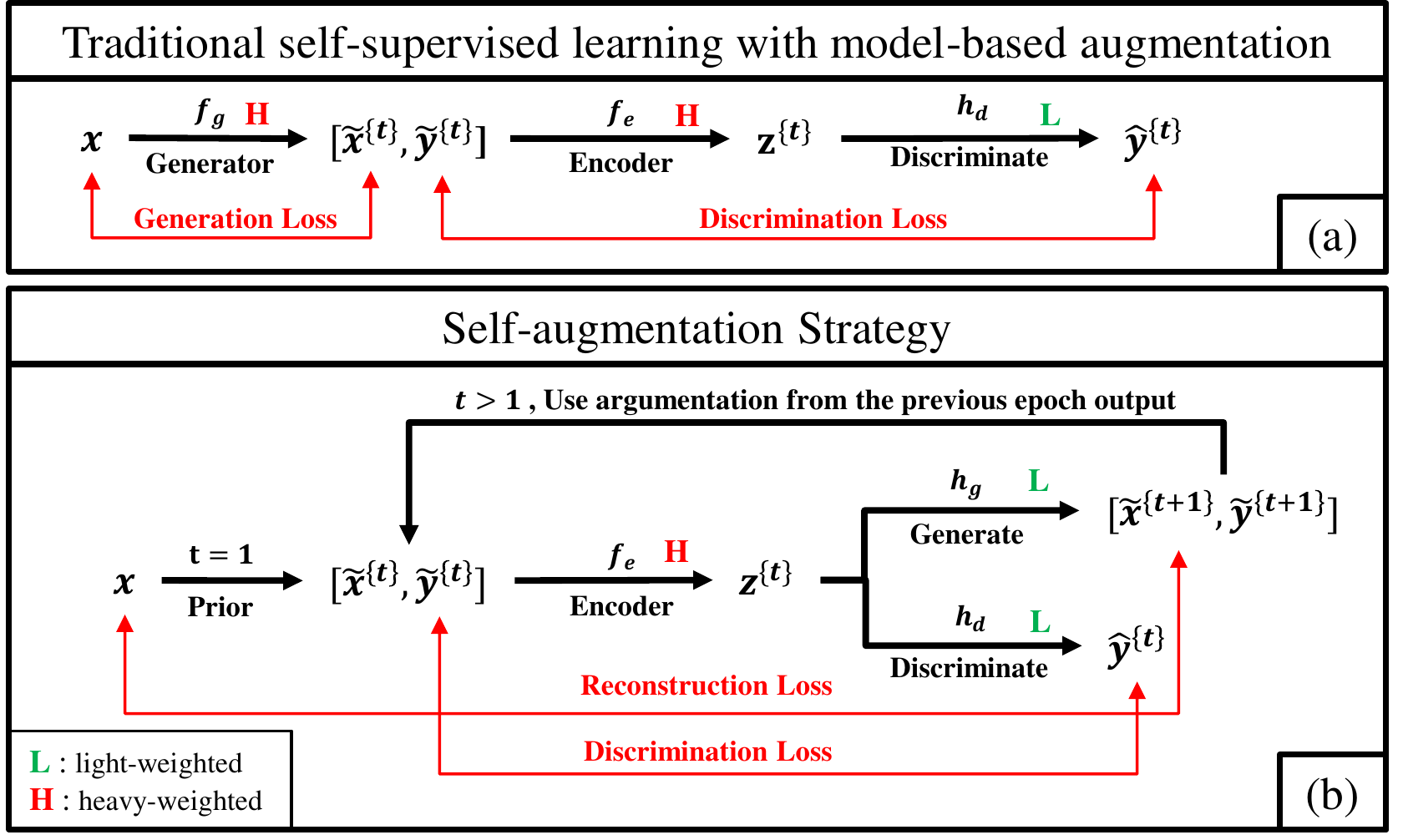} 
	\caption{The comparison between the traditional self-supervised learning with model-based augmentation and the self-augmentation strategy.}\label{fig:flow}
\end{figure}

In supervised learning, 
each data instance has both feature $\bm{x}_i$ and its corresponding label $\bm{y}_i$.
In self-supervised learning
such as language model pre-training,
since there is no label in the training dataset,
an input pair of $\bm{\tilde{x}}_i$ and label $\bm{\tilde{y}}_i$
is constructed by an augmentation process $q$ from the original data instance of token sequence $\bm{x}_i \sim \mathcal{X}$.
The augmentation process can be 
a context-independent process (such as random masking with a certain distribution) or a generation network $f_g$ with parameter $\theta_g$ which takes the sequence context into account.
Let's consider the latter and denote the augmented dataset as $ \tilde{{\mathcal{X}}} = \{[\bm{\tilde{x}}_i, \bm{\tilde{y}}_i] = f_g(\bm{x}_i) | \bm{x}_i \sim \mathcal{X}; i\in \{1,\cdots,n\}\}$.
As shown in Figure \ref{fig:flow}(a), the augmented data instance is fed to a heavy-weight main network with encoder $f_e$ with parameter $\theta_e$ and a light-weight head $h_d$ with parameter $\theta_d$ to conduct pre-training task. 
In a forward pass, the auxiliary generator and the main network are used sequentially. 
The auxiliary generator $f_g$ is usually a heavy-weighted network.
 
In SAS, we eliminate the separate generator in augmentation. Instead, we introduce both generation and discrimination capabilities via a single network with both MLM and RTD heads, $h_g$ and $h_d$, and design a light-weight generation process by using the softmax output of MLM. That is, the MLM head serves dual purposes for both pre-training and contextualized data augmentation. 
In the first epoch $(t=1)$,  
the augmented data is generated by a (cold-start) prior distribution $p^{\{0\}}$, since there is no prior MLM output yet.
In the $(t+1)$'th epoch for $t \geq 1$, 
the augmented data is generated based on MLM's softmax output of the previous epoch. 
During the training, while $h_g$  is updated from generation loss and $h_d$ from discrimination loss,  
the encoder $f_e$ is updated from both losses. 

\subsection{SAS in Language Model Pre-training}
Now we describe the details of SAS in the context of language model pre-training. 
As mentioned above, SAS has a single neural network that includes two light-weight heads
\begin{figure}
	\centering 
	\includegraphics[width=\linewidth]{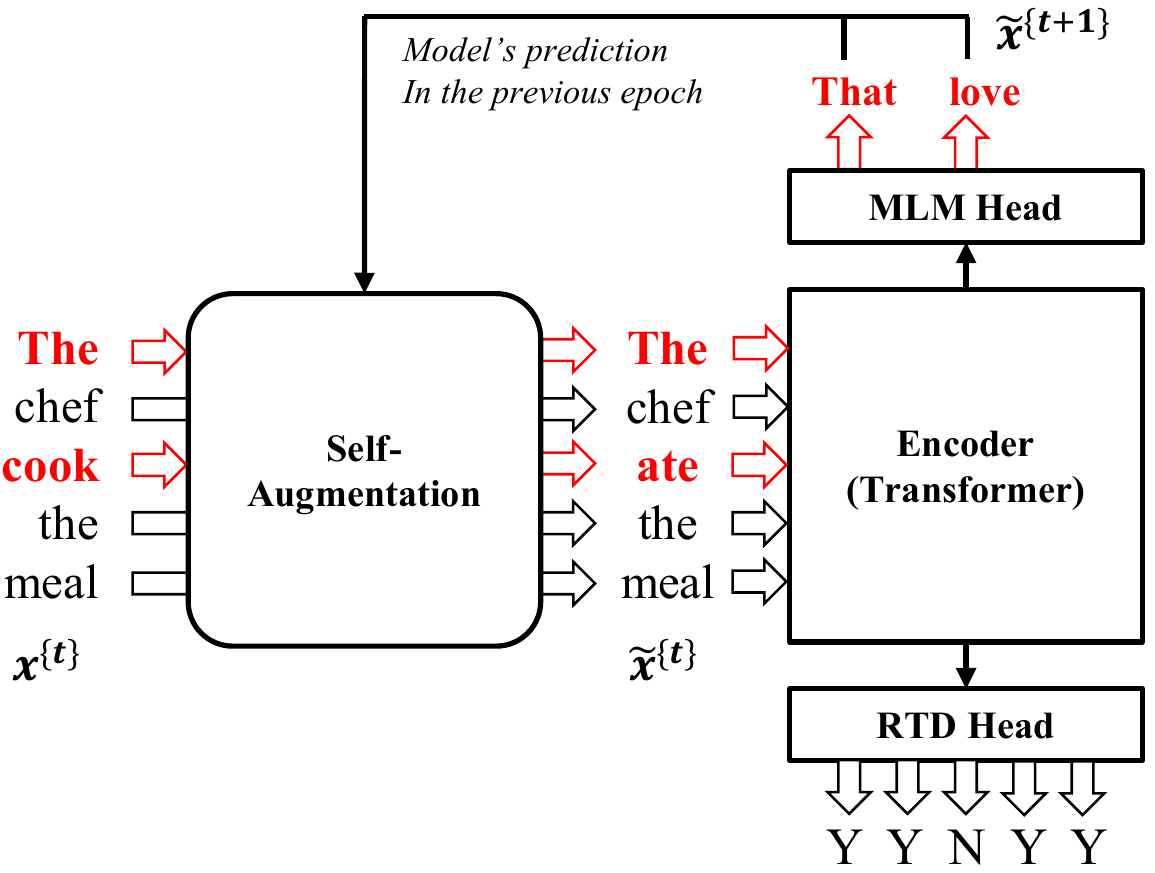} 
	\caption{Workflow of SAS in language model pre-training.}\label{fig:structure}
\end{figure}
on the top of one common heavy-weight transformer encoder. 
It undertakes both the MLM and RTD tasks in each forward pass during the training process. 
A workflow of SAS is demonstrated in Figure \ref{fig:structure}.

For an instance $\bm{x} = (x_1, x_2, \cdots, x_k)$ of a token sequence of length $k$ in the pre-training dataset $\mathcal{X}$, 
in the $t$'th epoch of the pre-training process,
we choose a random set $S^{\{t\}}$ with $\lceil 0.15 \cdot k \rceil$ position indexes, just as BERT and ELECTRA,
and create augmented input instance
$\bm{\tilde{x}}^{\{t\}} = (\tilde{x}_1^{\{t\}}, \tilde{x}_2^{\{t\}}, \cdots, \tilde{x}_k^{\{t\}} )$ as follows,
\begin{equation}\label{eq:z}
\left\{
\begin{array}{ll}
\tilde{x}_i^{\{t\}} = x_i           & {i \notin S^{\{t\}}}\\
\tilde{x}_i^{\{t\}} \sim p^{\{t-1\}}   & {i \in S^{\{t\}}}
\end{array} \right.
\end{equation}
In other words, we keep original token $x_i$ if $i \notin S^{\{t\}}$ 
and otherwise replace it by sampling a token based on a generating distribution $p^{\{t-1\}}$ to be detailed below. 

In the first epoch ($t=1$), we set $p^{\{0\}}$ 
as a cold-start token generating distribution, 
such as a (token-frequency based) unigram distribution\footnote{Our experimental results (in the appendix) show that the unigram
distribution and the uniform distribution work equally well for our SAS.}. 
Thereafter, in epoch $t+1$ ($t\geq 1$), we leverage the softmax output $p^{\{t\}}_{\theta_e,\theta_g}(x_i | \bm{\tilde{x}}^{\{t\}})$ of the MLM head in the $t$'th epoch as a contextualized token generating distribution. We have
\begin{equation}\label{eq_full_softmax}
\begin{split} 
    & p^{\{t\}}_{\theta_e,\theta_g}(x_i | \bm{\tilde{x}}^{\{t\}}) = \frac{ \exp \{ e(x_i)^T h_g( f_e( \bm{\tilde{x}}^{\{t\}} ) )_i \} } { \sum_{x' \in V} \exp \{ e(x')^T h_g( f_e( \bm{\tilde{x}}^{\{t\}} ) )_i \} },
\end{split}
\end{equation}
where $e(x_i)$ is the embedding of token $x_i$,
$f_e$ is the encoder that outputs a sequence of $k$ contextualized representation vectors from the augmented input $\bm{\tilde{x}}$, 
$h_g$ is the MLM head that generates representation vector at each augmented position in $S^{\{t\}}$, 
and $V$ is the vocabulary. 
Essentially, the SAS model itself in the $t$'th  epoch serves as a teacher to provide data augmentation for its learning in the $(t+1)$'th  epoch.

In practice, with the distribution $p^{\{t\}}_{\theta_e,\theta_g}(x_i | \bm{\tilde{x}}^{\{t\}})$ generated from the MLM head in the $t$'th epoch, we directly sample $\tilde{x}_i^{\{t+1\}}$ from it for every $i \in S^{\{t+1\}}$ and then store the sampled token index in CPU memory for the augmentation usage in the $(t+1)$'th epoch. This way, we avoid storing the whole distribution with high memory cost. The loss of the MLM task is calculated as a softmax loss,
\begin{equation}\label{eq_MLM_loss}
 \mathcal{L}_{mlm}^{\{t\}}(\bm{x}; \theta_e, \theta_g) = - \mathbb{E} \left[\sum_{i \in S^{\{t\}}} \log p^{\{t\}}_{\theta_e,\theta_g}(x_i | \bm{\tilde{x}}^{\{t\}}) \right].
\end{equation}

We adopt the same RTD task as ELECTRA to classify whether each token in the augmented input $\bm{\tilde{x}}^{\{t\}}$ is a replaced token or not. 
The corresponding classification probability is
\begin{equation}\label{eq_RTD_prob}
D(\bm{\tilde{x}}^{\{t\}})_i = \sigma(h_d(f_e(\bm{\tilde{x}}^{\{t\}}))_i ),
\end{equation}
where  $\sigma$ is the logistic sigmoid, 
$h_d$ is the RTD head function that generates a sequence of logits at all the %token
positions. Note that the same $f_e$ is used in Eq.\;(\ref{eq_full_softmax}) and Eq.\;(\ref{eq_RTD_prob}), indicating that only one forward pass through the encoder is needed. This largely reduces the computation cost, in contrast to ELECTRA's separate forward passes in two neutral networks.

Eq.\;(\ref{eq_RTD_prob}) leads to the loss of the RTD task given the instance  $\bm{x}$ in the $t$'th epoch as
\begin{equation}\label{eq_RTD_loss}
\begin{split}
    & \mathcal{L}_{rtd}^{\{t\}}( \bm{x}; \theta_e, \theta_d ) = - \mathbb{E} \bigg[\sum_{j = 1}^k y^{\{t\}}_j \log D(\bm{\tilde{x}}^{\{t\}})_j \\
    &+ (1-y^{\{t\}}_j) \log ( 1 - D(\bm{\tilde{x}}^{\{t\}})_j)  \bigg],
\end{split}
\end{equation}
where $y^{\{t\}}_j = 1$ if $\tilde{x}_j^{\{t\}} = x_j$, $y^{\{t\}}_j = 0$ otherwise.

We train the SAS model by minimizing a combined loss: 
\begin{equation}\label{eq_total_loss}
\begin{split}
\sum_{\bm{x} \in \mathcal{X}} \mathcal{L}_{mlm}^{\{t\}}(\bm{x}; \theta_e, \theta_g) +
   &  \lambda^{\{t\}} \mathcal{L}_{rtd}^{\{t\}}( \bm{x}; \theta_e, \theta_d ), 
\end{split}
\end{equation}
where $\lambda^{\{t\}}$ is the (relative) weight of the RTD loss  which can be a constant or varied across different epochs.
The joint-task learning with the combines loss, along with the shared encoder for both heads, avoids learning obstacles from potentially over-challenging replaced tokens.

\section{Experiments} \label{sec_expr}
In this section, we describe the experimental setup and corresponding experimental results.

\subsection{Experimental Setup}
\textbf{Pre-training Details:} 
Our implementation\footnote{Code and pretrained model is available publicly at Github: \url{https://github.com/alibaba/self-augmentation-strategy}.} is based on Huggingface Transformers 4.3 framework \citep{Transformers_Wolf_EMNLP2020}. 
We include ELECTRA, DeBERTa  as well as BERT for comparison. 
Under the current constraints of computation resource, we focus on the small and base models which have been extensively studied and compared by \citet{electra_clark2020}, and we set architectures and hyperparameters aligned with ELECTRA.
Please refer to the appendix for the detailed model architecture and pre-training hyperparameter values. %
We implement each model by largely re-using the corresponding code from Huggingface \citep{Transformers_Wolf_EMNLP2020},
if a pre-trained checkpoint has not been publicly released by its authors. We use the same pre-training data as BERT, ELECTRA-Small and ELECTRA-Base, which consists of 3.3 Billion tokens from Wikipedia and BooksCorpus datasets. 
For fair comparison, we follow \citet{electra_clark2020} to use FLOPs (floating point operations) to measure computation usage
(since FLOPs is a measure agnostic to the particular hardware and low-level optimizations).
We reuse the FLOPs computation code\footnote{See \url{https://github.com/google-research/electra/blob/master/flops_computation.py}} released from \citet{electra_clark2020} 
so that we essentially take the exactly same assumptions made by \citet{electra_clark2020}.
Some details of the experimented models are as follows.

\begin{itemize}
\item \textbf{ELECTRA}:
We train ELECTRA-Base and ELECTRA-Small using the exactly same hyperparameter values as \citet{electra_clark2020}, except for larger batch size and learning rate for ELECTRA-Small to reduce the pre-training time (which is not reflected in the FLOPs calculation). 
For ELECTRA-Small model as well as all other small models, we use batch size 512 and 0.25M pre-training steps, instead of batch size 128 and 1M steps in \citet{electra_clark2020}, and double the learning rate accordingly.\footnote{We observe that such a change is able to significantly reduce the pre-training time without degrading the model performance.} As a reference point, we also include ELECTRA-Small++ whose pre-trained model checkpoint is publicly released by \citet{electra_clark2020}.
Note that ELECTRA-Small++ uses 18x training FLOPs compared to ELECTRA-Small because it is pre-trained much longer with much larger data and its input sequence length is also quadrupled \citep{electra_clark2020}.

\item \textbf{DeBERTa}:
We implement DeBERTa and adopt its idea of sharing the projection matrices between relative position embedding and content embedding in 
all the attention layers proposed by \citet{deberta_he2020}.
Such a sharing does retain the model performance on the small and base models while reducing the number of model parameters.

\item \textbf{BERT}:
For BERT-Base, we use its model checkpoint publicly released by \citet{bert_devlin2018}.
We implement our BERT-Small and set its embedding size the same as its hidden size\footnote{\citet{electra_clark2020} define a different BERT-Small setting where its embedding size is decreased to half of its hidden size.
}, according to the convention of the BERT models.
Please refer to the appendix for the details about the hyperparameters.
Our BERT-Small setting makes its FLOPs similar to that of ELECTRA-Small when the training steps are the same, 
so that a fair comparison of their performance can be made directly.

\item \textbf{SAS}: We experiment three SAS settings: 

    \begin{itemize}
        \item 
        %SAS-(Small/Base) 
        SAS denotes the setting with a default weight scheduler that epoch-wisely increases
        the RTD loss weight $\lambda^{\{t\}}$ from 50 to 200 during the pre-training.
        \item SAS$^{c}$ denotes a setting where RTD loss weight $\lambda^{\{t\}}$ is a constant value 50.
        \item SAS$^{\text{DA}}$ denotes a setting that uses the default weight scheduler and further incorporates the disentangled attention mechanism.
    \end{itemize}
\end{itemize}
In all the SAS settings, 
we set the (token-frequency based) unigram distribution as the cold-start distribution.

\textbf{Downstream Tasks:} We evaluate all models on the General Language Understanding Evaluation (GLUE) benchmark \citep{GLUE_Wang_2018}. 
It contains a variety of tasks covering 
natural language inference tasks MNLI \citep{williams2017broad}, RTE \citep{giampiccolo2007third}, and QNLI \citep{rajpurkar2016squad};
semantic similarity tasks MRPC \citep{automatically_dolan2005}, QQP \citep{iyer2017first}, and STS-B \citep{semeval_cer2017};
sentiment classification task SST-2 \citep{recursive_socher2013};
and 
linguistic acceptability classification CoLA \citep{neural_warstadt2019}. See the supplementary for more details on the GLUE tasks. 

The evaluation metrics are 
the average of Spearman correlation and Pearson correlation for STS-B,
Matthews correlation for CoLA, the average of MNLI-match accuracy and MNLI-mismatch accuracy for MNLI, and accuracy for other GLUE tasks. 
We also take the average of metrics of these eight GLUE tasks, denoted by GLUE Score,  as the overall performance metric on these tasks.
All GLUE scores are based on the Dev dataset. 

\textbf{Fine-tuning Procedure:}
%fine-tuning
For the fine-tuning of GLUE, we add simple linear classifiers on top of the encoder of a pre-trained model. 
Because we observe a large performance variance in the GLUE tasks with small data sizes (such as CoLA, MRPC, STS-B and RTE), we adopt the following two methods to reduce the variance. 
First, we follow the strategy proposed in the papers \citep{stability_mosbach2020,revisiting_zhang2020,fine_dodge2020} to train more epochs with small learning rates for these small tasks.
Second, we fine-tune these small tasks by using multiple random seeds and obtain the average score across the seeds. 
Please refer to the supplementary for detailed fine-tuning hyperparameter settings.

%GLUE Score
For base models, we pre-train each model once and then use the above fine-tuning strategy to obtain the score of each GLUE task.
Since for some small models we still observe non-negligible variance of the resulting scores, we pre-train each small model using five different random seeds. The finally reported score of each task is the average across the five pre-trained model checkpoints.

\subsection{Overall Comparison Results}

\begin{table*}[t]
    \begin{tabular}{lp{1.3cm}p{1.5cm}cccccccc}
        \hline
        Model    					& Train FLOPs	& GLUE \newline mean\small{$\pm$std}					& CoLA & SST-2 & MRPC & STS-B & QQP & MNLI & QNLI & RTE \\
        \hline                                                                             
    BERT-Small   				& 1.274e18 		& 79.10\small{$\pm$0.07} 		& 49.69 & 90.14 & 84.64 & 86.04 & 89.51 & 80.00 & 86.84 & 65.94  \\
    BERT-Small-1.5x 			& 1.911e18			& 79.55\small{$\pm$0.09}  		& 51.80 	& 90.37 	& 84.07 & 86.04 & 89.80 & 80.20 & 86.58 & 67.51 \\
    ELECTRA-Small 			& 1.294e18 		& 80.78\small{$\pm$0.18} 		& 59.27	& 89.18	& 86.40	& 86.76	& 89.92	& 80.29 &	88.23 &66.21 \\
    \textit{ELECTRA-Small++}* 		& \textit{2.403e19}		& \textit{82.05} 						& \textit{58.37} & \textit{91.40} & \textit{87.01} & \textit{87.95} & \textit{90.54} & \textit{82.52} & \textit{88.93} & \textit{69.68}  \\
    DeBERTa-Small  			& 1.381e18  		& 79.52\small{$\pm$0.43} 		& 49.51 & 89.91 & 86.68 & 86.29 & 90.26 & 81.51 & 87.78 & 64.26  \\
        \hline
    SAS$^{c}$-Small  			& 1.279e18 		& 81.30\small{$\pm$0.13} 		& 59.52 & 89.60 & 87.17 & 87.27 & 90.20 & 81.54 & 88.67 & 66.43  \\
    SAS-Small  					& 1.279e18 		& 81.61\small{$\pm$0.24}			& 60.49 & 90.08 & 87.01 & 87.32 & 90.11 & 81.37 & 88.47 & \textbf{68.05} \\
    SAS$^{\text{DA}}$-Small  & 1.385e18 	& \textbf{82.14\small{$\pm$0.22}} 		& \textbf{62.35} & \textbf{90.55} & \textbf{87.55} & \textbf{87.52} & \textbf{90.60} & \textbf{82.20} & \textbf{88.71} & 67.65 \\
        \hline
    \end{tabular}
    \footnotesize{*: ELECTRA-Small++ is the pre-trained model publicly released by \citet{electra_clark2020}.  It uses 18x training FLOPs compared to ELECTRA-Small.}
    \caption{Comparison of small models on the GLUE dev set. }
    \label{tbl_small}
\end{table*}

\begin{table*}[h]
    \begin{tabular}{p{3.4cm}p{1.5cm}ccccccccc}
        \hline
        Model          						& Train FLOPs	& GLUE& CoLA & SST-2 & MRPC & STS-B & QQP & MNLI & QNLI & RTE \\
        \hline
        \textit{BERT-Base \small{(1M)}}*		& \textit{6.430e19} 		& \textit{83.06} 			& \textit{60.07} & \textit{92.09} & \textit{85.29} & \textit{89.22} & \textit{91.27} & \textit{83.99} & \textit{91.43} & \textit{71.12} \\
        ELECTRA-Base  	\small{(766K)}		& 6.426e19 		& 85.46     & 65.53 & 91.28 & \textbf{89.95} & \textbf{90.33} & 91.65 & 85.49 & 91.85 & \textbf{77.62}  \\
        DeBERTa-Base \small{(1M)}	& 7.443e19 		& 83.97 			& 58.46 & 93.23 & 88.97 & 89.36 & 91.37 & 85.53 & 91.52 & 73.29  \\
        SAS-Base  \small{(959K)}    & 6.196e19      & 85.20 	        & 66.24	& 92.20	& 89.22	& 90.30	& 91.46	& 85.54	& \textbf{91.95}	& 74.73 \\
        SAS$^{\text{DA}}$-Base \small{(833K)}    	& \textbf{6.226e19} 		& \textbf{85.60} 	& \textbf{66.56} & \textbf{93.35} & 88.73 & 90.05 & \textbf{91.73} & \textbf{86.49} & 91.74 & 76.17  \\
        \hline       
    \end{tabular}
    % }
    \footnotesize{*: BERT-Base is the model publicly released by \citet{bert_devlin2018}.}
    \caption{Comparison of base models on the GLUE dev set.
    }
    \label{tbl_base}
\end{table*}

The performance comparison among the small models is shown in Table \ref{tbl_small}.
In the table, the second column shows the training FLOPs of each model, 
and the third column lists the mean and the standard deviation of GLUE Score for each model across five independently pre-trained checkpoints.

As for the three competing models,
it shows that both ELECTRA-Small and DeBERTa-Small outperform BERT-Small in terms of GLUE Score.
Even when we increase the pre-training steps of BERT-Small model from 250K to 375K, denoted by BERT-Small-1.5x in the table, 
it still is worse than ELECTRA-Small while being similar to DeBERTa-Small.
The GLUE Score of ELECTRA-Small (our implementation) is about 98.45\% of that of ELECTRA-Small++ released by \citet{electra_clark2020} (80.78 vs. 82.05),
which is higher than the 97.87\% in Table 8 of the original paper \citet{electra_clark2020}. This verifies the correctness of our ELECTRA implementation.
Note that DeBERTa-Small has much higher MNLI mean score than ELECTRA-Small, while ELECTRA-Small has higher mean of GLUE Score which is aligned with the overall advantage of ELECTRA in small models emphasized by \citet{electra_clark2020}.

The proposed SAS model further improves the overall performance over ELECTRA and DeBERTa with fewer FLOPs.\footnote{With 1 V100 GPU, the pre-training of SAS$^{\text{DA}}$-Small takes 37.5h; both SAS-Small and SAS$^{\text{c}}$-Small takes about 24h;
and ELECTRA-Small takes about 35h.
} 
The mean of GLUE Score of SAS$^{c}$-Small (with the constant weight for the RTD loss) is 
0.52 point higher than that of ELECTRA-Small, while SAS with the epoch-wise weight increasing strategy  is 0.83 point higher.
As for MNLI task, both SAS$^{c}$-Small and SAS have their mean scores more than 1 point higher than ELECTRA.

Notably, SAS$^{\text{DA}}$-Small, the SAS with the disentangled attention mechanism, achieves a higher GLUE Score than ELECTRA-Small++'s with only 5.76\% of FLOPs consumed by ELECTRA-Small++. 
The disentangled attention mechanism embraced by SAS is able to improve the mean of GLUE Score by 0.65\% (from 81.61 to 82.14).
This is larger than its 0.53\% improvement ratio on the basis of BERT (from 79.10 to 79.52), 
which shows that SAS can effectively realize the value of the disentangled attention mechanism.
The table also shows that DeBERTa-Small 
has the largest standard deviation among all the models.
This might be because
DeBERTa-Small needs larger data (such as the data used by \citet{deberta_he2020}) to stably achieve its functionality.

The comparison results on the base models are shown in Table \ref{tbl_base}. 
In the first column of the table, we show the pre-training steps of each model and have ensured that SAS takes similar or fewer FLOPs than other models.
The table shows that SAS-Base has a slightly lower GLUE Score than ELECTRA-Base, while
SAS$^{\text{DA}}$-Base\footnote{
The pre-training costs 7.7 days by 8 V100 GPUs.} is able to achieve both the highest GLUE Score (ELECTRA-Base is a very close second) and the highest MNLI score (nearly 1 point higher than the second best DeBERTa-Base).

\subsection{Ablation Study}

We design the following variants to investigate the contributions of different components of SAS:

\begin{itemize}
\item Unig-MLM denotes a variant of BERT with random replacement of 15\% of tokens based on the (token-frequency based) unigram distribution. That is, each of the randomly selected tokens is replaced with a token sampled from the unigram distribution instead of a [MASK] token.

\item Unig-MLM-SAS denotes a variant of SAS with only MLM task instead of two combined tasks.

\item Unig-MLM-RTD$^c$ denotes a model that is the same as SAS$^c$ but use a random replacement with the unigram distribution in the whole training process.

\item Unig-MLM-RTD denotes a model that is the same as SAS but use a random replacement with the unigram distribution in the whole training process.

\end{itemize}

\begin{table*}[h]
    % \vspace{20pt}
    \centering
    % \resizebox{142mm}{!}{%
    \begin{tabular}{p{3.3cm}p{1.8cm}p{1cm}p{1cm}p{1cm}p{1.2cm}p{1cm}p{1cm}p{1cm}p{1cm}}
        \hline
        Model          			& GLUE \newline mean\small{$\pm$std}		& CoLA & SST-2 & MRPC & STS-B & QQP & MNLI & QNLI & RTE \\
        \hline
    Mask-MLM \small{(BERT)} & 79.10 \small{$\pm$0.12} & 49.69 & 90.14 & 84.64 & 86.04 & 89.51 & 80.00 & 86.84 & 65.94  \\    
    Unig-MLM     				& 79.00 \small{$\pm$0.07}	& 49.61 & 90.06 & 84.97 & 85.43 & 89.42 & 80.11 & 86.93 & 65.46  \\
    Unig-MLM-SAS       		& 79.38 \small{$\pm$0.36}	& 49.94 & 89.72 & 86.13 & 85.48 & 89.69 & 80.34 & 87.43 & 66.28  \\
    Unig-MLM-RTD$^c$    & 80.43 \small{$\pm$0.14}	& 59.28 & 89.37 & 85.46 & 85.47 & 89.87 & 80.50 & 88.17 & 65.34  \\
    Unig-MLM-RTD   		& 80.40 \small{$\pm$0.15}	& 59.96 & 89.53 & 85.05 & 85.36 & 89.76 & 80.56 & 87.97 & 64.98  \\
    SAS$^{c}$ 				& 81.30 \small{$\pm$0.13}	& 59.52 & 89.60 & 87.17 & 87.27 & 90.20 & 81.54 & 88.67 & 66.43  \\
    SAS 							& 81.61 \small{$\pm$0.24}		& 60.49 & 90.08 & 87.01 & 87.32 & 90.11 & 81.37 & 88.47 & 68.05 \\
        \hline
    \end{tabular}
    \caption{Ablation study on small models on the GLUE dev set. }
    \label{tbl_ablation}
\end{table*}

Table \ref{tbl_ablation} summarizes the results on the small models.  
First, it shows slight drop (0.1 point) in GLUE Score if the random masking of BERT is changed to the random replacement, which might be among the reasons that BERT uses the special [MASK] token despite its mismatch with the fine-tuning data.
Second, when we use SAS to conduct the MLM task alone, it only increases the mean of GLUE Score from 79.10 to 79.38 compared to BERT, which indicates that the MLM task cannot fully utilize the self-augmented data to learn a better model.
Third, by comparing Unig-MLM to Unig-MLM-RTD (or  Unig-MLM-RTD$^c$), 
we observe that a big (about 1.4 point) increase in the mean of GLUE Score comes from adding the RTD task with the MLM task.
This increase is mainly due to the improvement in CoLA, as it alone contributes more than a 1.2 point increase in the mean of GLUE Score. 
This indicates that the RTD task along with the data augmented from the unigram distribution can greatly help the syntactic task (CoLA) but not other semantic tasks in GLUE.
Finally, by comparing Unig-MLM-RTD (or  Unig-MLM-RTD$^c$) to SAS (or SAS$^{c}$),
we see another notable increase in the mean of GLUE Score, which comes from the improvement in both the syntactic task (CoLA)  and other semantic tasks such as MNLI. This shows that SAS can further improve a variety of different tasks, as its pre-trained model has both generation and discrimination capabilities with MLM and RTD heads together. 
The effect of the weight strategy for the RTD loss can also be seen from Table \ref{tbl_ablation}.
By comparing SAS$^{c}$ to SAS, 
we see that CoLA's score has a further increase when the epoch-wise weight increasing strategy gradually puts more emphasis on the RTD task.

\subsection{Pre-Training Efficiency}

\begin{figure}[!t]
	\centering 
	\includegraphics[width=\linewidth]{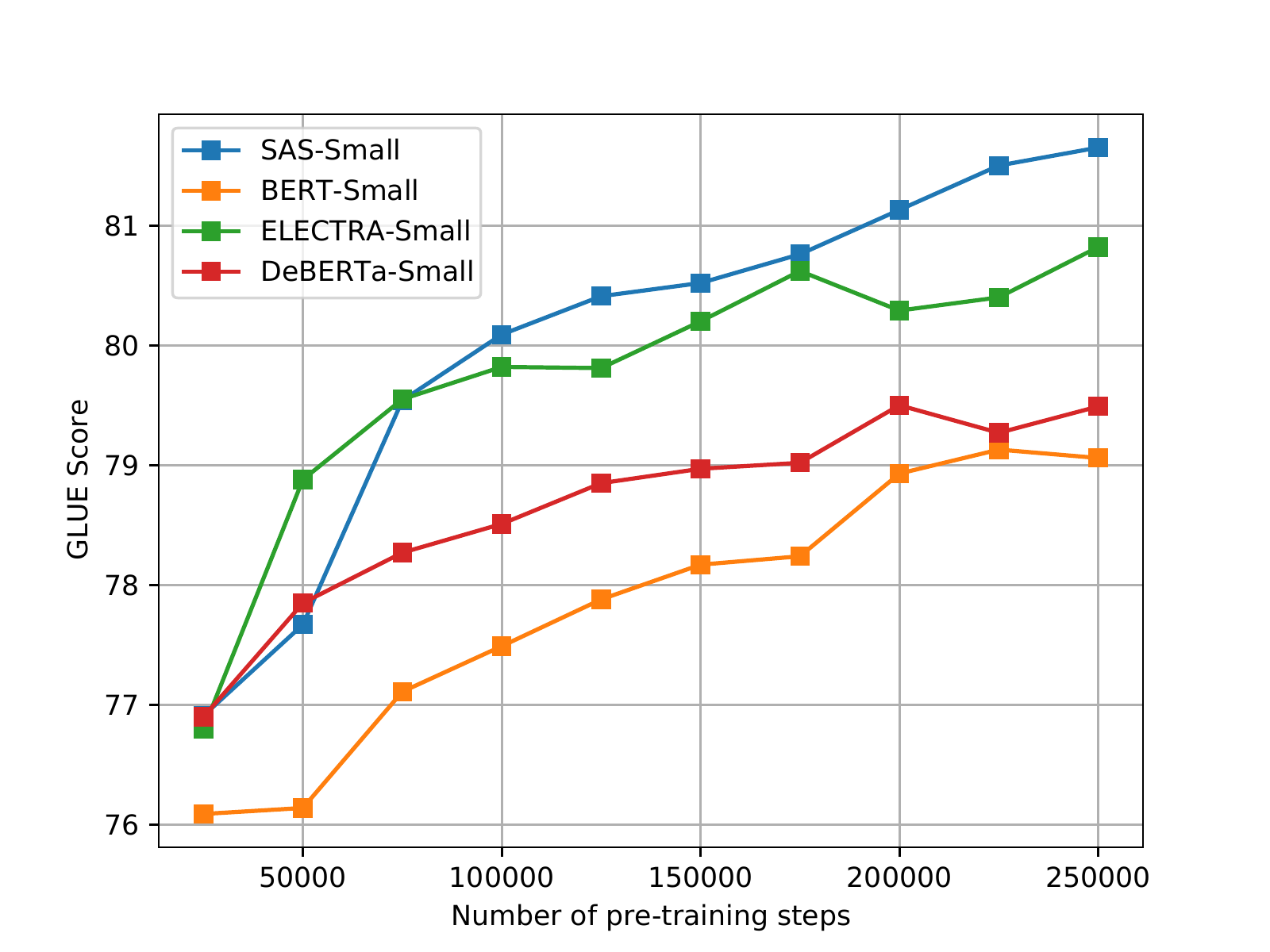} 
\caption{Curve of GLUE Score for SAS-Small model and its competitors with respect to the pre-training steps on the GLUE dev set.}
\label{fig_efficiency_score8}
\end{figure}

\begin{figure}[t]
	\centering 
	\includegraphics[width=\linewidth]{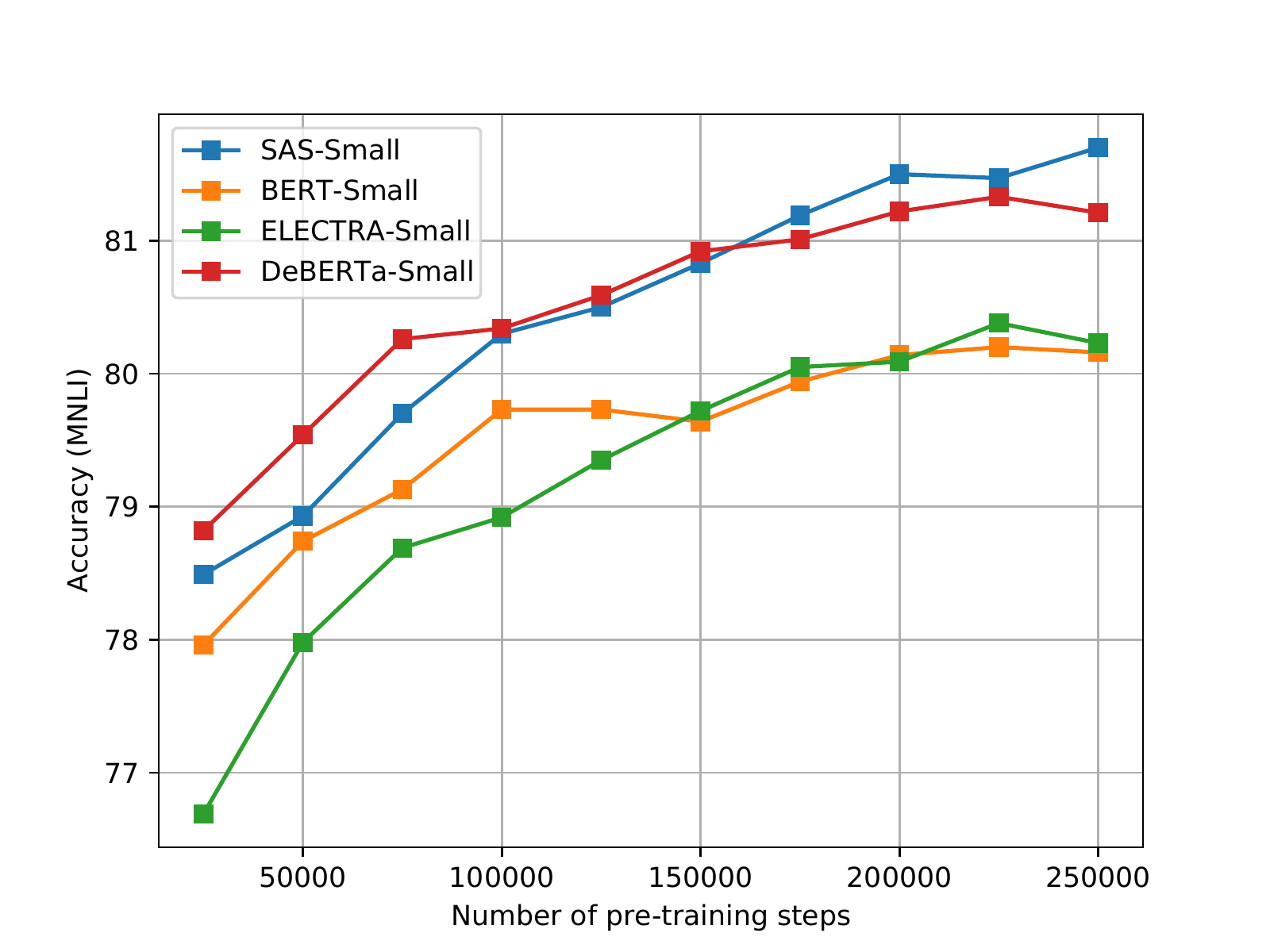} 
\caption{Curve of MNLI's average accuracy for SAS-Small model and its competitors with respect to the pre-training steps on the GLUE dev set.}
\label{fig_efficiency_MNLI}
\end{figure}

To investigate the convergence of pre-training, 
in Figures \ref{fig_efficiency_score8} and \ref{fig_efficiency_MNLI}, we plot GLUE Score and MNLI score with respect to the number of pre-training steps for 
SAS-Small and other models.
For each model, we select the median run whose pre-training random seed achieves the median GLUE Score among the five random seeds.
Then for the selected median run of each model, we save a checkpoint every 25K pre-training steps (i.e., 0.5 epoch),
and fine-tune it on each GLUE task and finally report GLUE Score across the tasks. 
In Figure \ref{fig_efficiency_score8}, it shows that
SAS-Small is significantly better than BERT-Small during the whole pre-training process, and starts to outperform all other models after the 1.5-th epoch. 
Figure \ref{fig_efficiency_MNLI} shows that SAS-Small and DeBERTa-Small perform consistently better than BERT-Small and ELECTRA-Small for MNLI task, while SAS-Small performs on a par with DeBERTa-Small since the second epoch.

\section{Conclusion} \label{sec_conclusion}
We propose the self-augmentation strategy (SAS) motivated %inspired 
by ELECTRA's contextualized data augmentation for language model pre-trainings. 
SAS eliminates a separate generation network used in ELECTRA and uses a single network to generate the contextualized data augmentation for its own subsequent training. 
By this way, it avoids the challenge in controlling the relative capability of the generator and reduces the computation cost.
In addition, our SAS is a general and open framework so that it can easily incorporate many other techniques emerging recently or in the future to further improve its performance.
The experimental results show that
SAS is able to outperform the previous state-of-the-art language pre-training models in the small and base scales with similar or less computation cost.
We will continue to validate SAS's advantage in
larger models when sufficient computation resources are available, while encouraging more studies on economic and compute-efficient pre-training approaches.

\bibliography{aaai22}

\clearpage

\begin{table*}[!b]
    \centering
    \begin{tabular}{lp{1.9cm}cccccccc}
        \hline
        Model    					& GLUE Score \newline mean\small{$\pm$std}			& CoLA & SST-2 & MRPC & STS-B & QQP & MNLI & QNLI & RTE \\
        \hline                                                                             
    Unig-SAS$^{c}$-Small  			& 81.30$\pm$0.13 		& 59.52 & 89.60 & 87.17 & 87.27 & 90.20 & 81.54 & 88.67 & 66.43  \\
    Unig-SAS-Small  					& 81.61$\pm$0.03			& 60.49 & 90.08 & 87.01 & 87.32 & 90.11 & 81.37 & 88.47 & 68.05 \\
    Unif-SAS$^{c}$-Small  			& 81.49$\pm$0.17 		& 59.59	 & 89.98 & 87.35 &87.39 & 90.15 & 81.44 & 88.26 & 67.73  \\
    Unif-SAS-Small  					& 81.57$\pm$0.15			& 60.59 & 90.34 & 87.21 & 87.35 & 90.10 & 81.39 & 88.09 & 67.51	 \\
        \hline
    \end{tabular}
    \caption{Comparison of cold-start token generating distributions of SAS-Small on the GLUE dev set. 
    }
    \label{tbl_cs_gene_distn}
\end{table*}

\section{Appendices}

\subsection{Study on Cold-start Token Generating Distribution}
We experiment with the two common choices for the cold-start token generating distribution for our SAS: 
the uniform distribution and the (frequency-based) unigram distribution.
The results are shown in Table \ref{tbl_cs_gene_distn}. 
We use the prefixes "Unig" and "Unif" to denote the unigram distribution and the uniform distribution respectively.
The table clearly shows that there is no statistically significant difference in the GLUE average scores between the two distribution choices. They work equally well for our SAS.

\subsection{Pre-training Details}
The following pre-training details apply to our SAS and its competing methods include the BERT, the ELECTRA and the DeBERTa. 
We always use Adam as the optimizer with weight decay. 
We mostly use the same hyperparameters as BERT and ELECTRA. 
For our SAS, we introduce a scheduler for $\lambda^{\{t\}}$, the (relative) weight for the RTD loss in Eq. 6,
which has an epoch-wise linear increase from 50 to 200 by default. 
We use the whole-word and dynamic token masking with the masked positions decided on the fly instead of during the preprocessing. 
Our own implementation also does not include the next sentence prediction (NSP) task proposed in the original BERT, 
as the recent works such as \citet{RoBERTa_Liu_2019} have suggested that it does not improve the performance. 
For the investigated DeBERTa baseline, we use the same corpus, tokenizer, training steps as the BERT for a fair comparison.
We searched for the best learning rate for Small models out of [1e-3, 5e-4] and selected $\lambda$ out of [1, 25, 50, 100] in early experiments. 
Otherwise, we did no hyperparameter tuning beyond the experiments. 
The full set of hyperparameters is listed in Table \ref{tbl_pretraing-details}.

\begin{table*}[h]
    % \vspace{20pt}
    \centering
    \begin{tabular}{p{4cm}p{2.8cm}p{2.5cm}p{2.8cm}}
        \hline
        Hyperparameter          	& ELECTRA-Small 	& All Other Small Models	& All Base Models\\
        \hline
        Number of layers     		& 12 			& 12 			& 12  \\
        Hidden size          			& 256 		& 256 		& 768 \\
        FFN inner hidden size 	& 1024 		& 1024 		& 3072 \\
        Attention heads       		& 4 			& 4 			& 12 \\
        Attention head size   		& 64  		& 64  		& 64  \\
        Embedding size        		& 128 		& 256 		& 768  \\
        Sequence length      		& 128 		& 128 		& 512  \\
        Mask percent         			& 15   		& 15   		& 15   \\
        Learning rate decay   	& Linear 	& Linear 	& Linear \\
        Warmup steps				& 10000	& 10000	& 10000 \\
        Learning rate      			& 1e-3    	& 1e-3    	&  2e-4  \\
        Adam $\epsilon$   			& 1e-6    	& 1e-6    	&  1e-6  \\
        Adam $\beta_1$    			&  0.9    	&  0.9    	&  0.9  \\
        Adam $\beta_2$    			&  0.999   &  0.999   	&  0.999 \\
        Attention dropout    		& 0.1  		& 0.1  		&  0.1  \\
        Dropout     						&   0.1    	&   0.1    	&   0.1  \\
        Weight decay   	 			&   0.01  	&   0.01  	&  0.01  \\
        Batch size   					&   512    	&   512    	&  256  \\
        Train steps  					&  0.25M   &  0.25M   & 766K - 1M \\
        \hline
    \end{tabular}
    \caption{Pre-training hyperparameters for all the models pre-trained by us.}
    \label{tbl_pretraing-details}
\end{table*}

\begin{table*}[h]
    % \vspace{20pt}
    \centering
    \begin{tabular}{ll}
        \hline
        Hyperparameter          & Value  \\
        \hline
        Learning rate     & 1e-4, 7.5e-5 for Small;  7.5e-5, 5e-5 for Base  \\     
        Adam $\epsilon$   &   1e-6           \\
        Adam $\beta_1, \beta_2$    &   0.9, 0.999         \\
        Layerwise LR decay   &  None \\
        Learning rate decay   &  Linear \\
        Warmup fraction    &   0.1  \\
        Attention dropout   &   0.1  \\
        Dropout     &    0.1  \\
        Weight decay    &  None \\
        Batch size   &   16, 32 \\
        Train epochs       & 20 for CoLA, MRPC, STS-B, RTE; 4 for other tasks \\
        Seeds          & 5 for CoLA, MRPC, STS-B, RTE; 3 for QNLI, SST2; 1 for MNLI, QQP \\
        \hline
    \end{tabular}
    \vspace{1mm}
    \caption{Fine-tuning hyperparameters for all the investigated models.}
    \label{finetune-details}
\end{table*}

\subsection{Fine-tuning Details}
We originally fine-tuned all the pre-trained models for 4 epochs. 
However, because we observed a large variance in the small tasks in GLUE, following the advice from \citet{stability_mosbach2020}, we increase the fine-tuning process to 20 epochs and select the best epoch for the four small tasks including CoLA, MRPC, STS-B and RTE. 
For Small models, we searched for the best learning rate out of [1e-4, 7.5e-5].
For Base models, we searched for a learning rate out of [5e-5, 7.5e-5] without the layer-wise learning-rate decay proposed by ELECTRA, but otherwise used the same hyperparameters as for small models. 
Due to limited computation resource, 
we adjust the number of independent fine-tuning runs (with different random seeds) so that we fine-tune more times for these tasks with smaller data sizes (i.e., with more variability).
The full set of hyperparameters is listed in Table \ref{finetune-details}. Following the BERT and the ELECTRA, we do not show results on the WNLI GLUE task for the Dev set results.

\subsection{Details about GLUE}
We provide further details about the GLUE benchmark tasks as follows.

\textbf{CoLA}: Corpus of Linguistic Acceptability \citep{neural_warstadt2019}. The task is to determine whether a given sentence is grammatical or not. The dataset contains 8.5k train examples from books and journal articles on linguistic theory.

\textbf{SST-2}: Stanford Sentiment Treebank \citep{recursive_socher2013}. The task is to determine if the sentence is positive or negative in sentiment. The dataset contains 67k train examples from movie reviews.

\textbf{MRPC}: Microsoft Research Paraphrase Corpus \citep{automatically_dolan2005}. The task is to predict whether two sentences are semantically equivalent or not. The dataset contains 3.7k train examples from online news sources.

\textbf{STS-B}: Semantic Textual Similarity \citep{semeval_cer2017}. The task is to predict how semantically similar two sentences are on a 1-5 scale. The dataset contains 5.8k train examples drawn from news headlines, video and image captions, and natural language inference data.

\textbf{QQP}: Quora Question Pairs \citep{iyer2017first}. The task is to determine whether a pair of questions are semantically equivalent. The dataset contains 364k train examples from the community question-answering website Quora.

\textbf{MNLI}: Multi-genre Natural Language Inference \citep{williams2017broad}. Given a premise sentence and a hypothesis sentence, the task is to predict whether the premise entails the hypothesis, contradicts the hypothesis, or neither. The dataset contains 393k train examples drawn from ten different sources.

\textbf{QNLI}: Question Natural Language Inference; constructed from SQuAD \citep{rajpurkar2016squad}. The task is to predict whether a context sentence contains the answer to a question sentence. The dataset contains 108k train examples from Wikipedia.

\textbf{RTE}: Recognizing Textual Entailment \citep{giampiccolo2007third}. Given a premise sentence and a hypothesis sentence, the task is to predict whether the premise entails the hypothesis or not. The dataset contains 2.5k train examples from a series of annual textual entailment challenges.
\end{document}